\documentclass[runningheads]{llncs}
\usepackage[T1]{fontenc}
\usepackage{graphicx}
\usepackage{graphicx}
\usepackage{amsmath}
\usepackage{subfig}
\usepackage{url}
\usepackage{soul}
\usepackage{hyperref} 
\usepackage{xspace}
\usepackage{algpseudocode}
\usepackage{algorithm}
\usepackage{multirow}
\usepackage[table,xcdraw]{xcolor}
\usepackage{orcidlink}
\usepackage{mathtools}

\usepackage{booktabs}
\makeatletter
\DeclareRobustCommand\onedot{\futurelet\@let@token\@onedot}
\def\@onedot{\ifx\@let@token.\else.\null\fi\xspace}

\def\eg{\emph{e.g}\onedot} 
\def\ie{\emph{i.e}\onedot}

\def\etal{\emph{et al}\onedot}
\makeatother

\algnewcommand{\LeftComment}[1]{\Statex \(\triangleright\) #1}

\newcommand{\blue}[1]{\textcolor{blue}{#1}}

\newcommand{\ourname}{VLTNet}

\begin{document}
\title{Zero-shot Object Navigation with Vision-Language Models Reasoning}
\author{Congcong Wen\inst{1,2} \and
Yisiyuan Huang\inst{3} \and
Hao Huang\inst{1,2} \and 
Yanjia Huang\inst{3} \and 
Shuaihang Yuan\inst{1,2} \and 
Yu Hao\inst{1,2} \and 
Hui Lin\inst{4} \and 
Yu-Shen Liu\inst{5} \and 
Yi Fang\inst{1,2}}

% \author{Congcong Wen\inst{1,2}\orcidID{0000-0001-6448-003X
% } \and
% Yisiyuan Huang\inst{3}\orcidID{1111-2222-3333-4444} \and
% Hao Huang\thanks{Corresponding Author.}\inst{1,2}\orcidID{0000-0002-9131-5854} \and 
% Yanjia Huang\inst{3}\orcidID{0009-0009-1092-3275} \and 
% Shuaihang Yuan\inst{1,2}\orcidID{0000-0002-7092-7966} \and 
% Yu Hao\inst{1,2}\orcidID{0000-0001-9119-6114} \and 
% Hui Lin\inst{4}\orcidID{0000-0003-0190-969X} \and 
% Yu-Shen Liu\inst{5}\orcidID{0000-0001-7305-1915} \\ \and 
% Yi Fang\inst{1,2}\orcidID{0000-0001-9427-3883}}
%
\authorrunning{C. Wen et al.}
\institute{Embodied AI and Robotics (AIR) Lab,
New York University Abu Dhabi, UAE \and
Center for Artificial Intelligence and Robotics, \\ New York University Abu Dhabi, UAE \and
Tandon School of Engineering, New York University, New York, USA. \and 
China Academic of Electronics and Information Technology, Beijing, P.R. China. \and
School of Software, Tsinghua University, Beijing, P.R. China. }
% \thanks{* indicates equal contribution.}
\maketitle              % typeset the header of the contribution
\begin{abstract}

Object navigation is crucial for robots, but traditional methods require substantial training data and cannot be generalized to unknown environments. Zero-shot object navigation (ZSON) aims to address this challenge, allowing robots to interact with unknown objects without specific training data. Language-driven zero-shot object navigation (L-ZSON) is an extension of ZSON that incorporates natural language instructions to guide robot navigation and interaction with objects. In this paper, we propose a novel Vision Language model with a Tree-of-thought Network (VLTNet) for L-ZSON. VLTNet comprises four main modules: vision language model understanding, semantic mapping, tree-of-thought reasoning and exploration, and goal identification. Among these modules, Tree-of-Thought (ToT) reasoning and exploration module serves as a core component, innovatively using the ToT reasoning framework for navigation frontier selection during robot exploration. Compared to conventional frontier selection without reasoning, navigation using ToT reasoning involves multi-path reasoning processes and backtracking when necessary, enabling globally informed decision-making with higher accuracy. Experimental results on PASTURE and RoboTHOR benchmarks demonstrate the outstanding performance of our model in LZSON, particularly in scenarios involving complex natural language as target instructions. 

% Videos are available at \url{https://vlt-lzson.github.io/}.

\keywords{Zero-shot Object Navigation  \and Vision-Language Model (VLM) \and Large Language Mdoel (LLM) \and LLM Reasoning.}
\end{abstract}

\section{Introduction}

Object navigation, a fundamental task in robotics, is crucial for robots to intelligently explore an environment and interact with objects in the environment. Conventional methods rely on extensive visual training data containing labeled objects from the environment, limiting their ability to generalize to unknown and unstructured environments. To remedy this limitation, recent research~\cite{Zhao_2023,Majumdar_2022,Dorbala_2022,Park_Yoon_023} explores zero-shot object navigation (ZSON), which allows robots to navigate and interact with unknown objects without the corresponding labeled training data. However, while effective in basic navigation, this method often falls short in scenarios requiring intricate interaction and communication, which are essential for enhanced autonomy and more robust human-robot collaborations. 

\begin{figure}[t]
    \centering
    \medskip
    \includegraphics[width=1.0\linewidth]{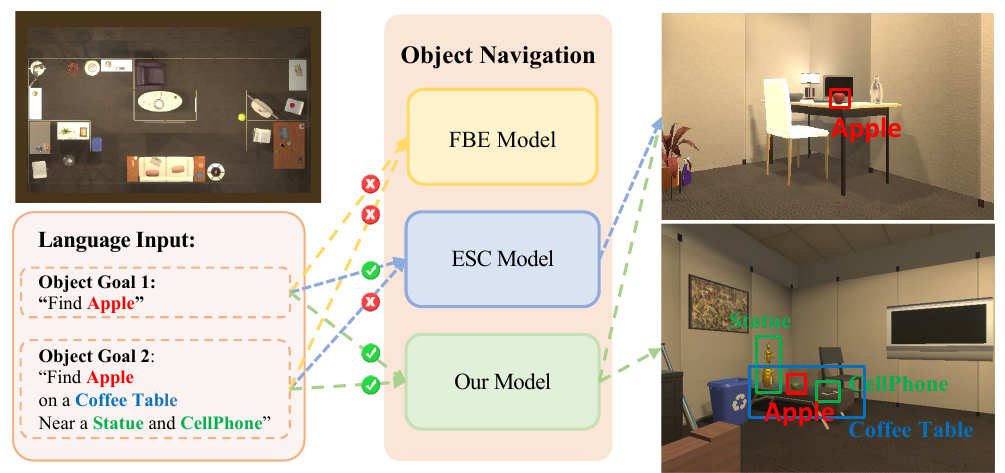}
    % \vspace{-5mm}
    \caption{Comparison of different object navigation methods under two types of language input: 1) word input with only object category, 2) sentence input with detailed spatial descriptions. a) FBE model~\cite{Yamauchi}: cannot accept either word or sentence input. b) ESC model~\cite{Zhou_2023}: only accepts word input. c) Our model: accepts both word and sentence as input.
    }
    % \vspace{-3mm}
    \label{fig:intro}
\end{figure}

To improve autonomous agents and human-robot interaction which is lacking in the traditional ZSON, there is a growing interest in Language-driven Zero-Shot Object Navigation (L-ZSON). L-ZSON guides agents using natural language instructions to require agents to follow the textual or spoken guidance to reach the specified unseen objects or locations. The pioneering efforts~\cite{Gadre_2023,Dorbala_2022,Park_Yoon_023} have leveraged Large Language Models (LLMs) for L-ZSON. For instance, Huang et al. \cite{huang2023visual} introduce VLMaps, a spatial map representation that integrates pre-trained visual-language features with 3D reconstruction of a physical environment. Zhou et al.~\cite{Zhou_2023} introduce a novel Exploration with Soft common sense Constraints (ESC) module that utilizes a pre-trained LLM for scene understanding and common sense reasoning. Nonetheless, these approaches can only handle instructions that explicitly contain object categories, failing to navigate to unknown objects or objects described by spatial or visual attributes in the instructions. To remedy the problem, Gadre et al.~\cite{Gadre_2023} build the PASTURE benchmark, which more closely reflects real-world scenarios and provides a more rigorous evaluation of L-ZSON.  Therefore, we choose this benchmark to evaluate the performance of our proposed L-ZSON method. Furthermore, it is worth noting that existing  works employ standard LLMs for common-sense reasoning or decision-making. However, although LLMs are powerful in many applications, they can still struggle to self-assess their decisions during reasoning processes, thus potentially leading to sub-optimal decisions.

To resolve this critical problem of making more effective decisions in dynamic environments, in this paper, we propose a  novel Vision Language Model with Tree-of-thoughts NETwork, named \textit{VLTNet}, for L-ƒZSON. VLTNet consists of four  core modules: vision language model understanding, semantic mapping, tree-of-thoughts reasoning and exploration, and goal identification. Specifically, we first leverage the vision language model understanding module to perform scene understanding. Then, we use the semantic mapping module to build a semantic navigation map. Next, we utilize the tree-of-thoughts reasoning and exploration module to select frontiers based on common sense reasoning for exploration. Finally, we employ the goal identification module to determine whether the current object being navigated to matches the target object. A significant novelty of our paper is the utilization of the Tree-of-Thoughts (ToT) reasoning framework for frontier selection in robot exploration. As shown in Fig. \ref{fig:intro}, our model with ToT reasoning can incorporate goal-based instructions of varying complexity to choose the optimal frontier. Unlike conventional LLMs, ToT equips models with the capacity to engage in deliberate, multi-path reasoning processes, enabling them to self-evaluate choices and make informed decisions for the action. This self-evaluation reasoning framework also allows models to anticipate future prediction and backtrack when necessary to make globally informed decisions. Experimental results conducted on two benchmarks, PASTURE \cite{Gadre_2023} and RoboTHOR \cite{Deitke_2020}, demonstrate that our model excels in L-ZSON tasks, particularly in complex ZSON tasks that involve natural language as guidance.

\section{Related Work}

\paragraph{Object Goal Navigation}
The primary task of goal-conditioned navigation is to guide robots towards distinct targets based on varying specifications. These specifications can be categorized into position goals, \ie, predefined spatial coordinates \cite{Chaplot,Chatto_2021}; image goals, \ie, locations that match a given image view \cite{Mezghan_2022,Zhu_2017}; and object goals, \ie, locations containing specific objects that the agent needs to find~\cite{Gadre_2023,Halah_2022,Zhou_2023,Chang_Gupta_Gupta_2020}.
Our research focus on object goal navigation task, which requires the robot to locate and navigate towards specific objects within an environment.

In order to develop agents capable of navigating previously unseen environments, recent work has shifted focus to Zero-shot Object Navigation (ZSON) \cite{Zhao_2023,Majumdar_2022,Dorbala_2022,Park_Yoon_023}. Nonetheless, most ZSON approaches only take in object names as targets, which can sometimes lead to inefficiency and inaccuracy when navigating through complex environments. Therefore, Language Driven Zero-shot Object Navigation (L-ZSON)  were studied as a subset of ZSON, aiming to interpret object goals and descriptive cues from natural language input \cite{Gadre_2023,Dorbala_2022,Park_Yoon_023}.

\paragraph{Exploration strategies}
Currently, the exploration strategies in object goal navigation can be divided into two main categories: learning-based and frontier-based.

\textbf{Learning-based exploration} strategies can be divided into two lines. The first utilizes pre-trained visual encoders \cite{He_2017,Radford_2021} to convert egocentric images into descriptive feature vectors, which were fed to train a robot navigation policy by employing imitation learning or reinforcement learning \cite{Ye_Batra_Das_Wijmans_2021,Maksymets_Cartillier_Gokaslan_Wijmans_Galuba_Lee_Batra_2021,Khandelwal_2022,Ramrakhya_2022,Deitke_2022,Chen_2022}. 
The second constructs an explicit semantic graph and then the navigation policies are then trained to identify locations of goal objects with the semantic graphs \cite{Chaplot,Min_2023,Zheng_2022}. 
Learning-based object goal navigation methods rely on training data to fine-tune the navigation policies of agents, often necessitating intricate reward engineering \cite{Gadre_2023}. Furthermore, these methods often face difficulties in generalizing with new objects or unfamiliar environments drastically different from their training data \cite{Zhou_2023}. 

\textbf{Frontier-based exploration} strategies address the limitations of the learning-based approaches. 
Frontier-based exploration (FBE) \cite{Yamauchi} is a heuristic algorithm to navigate a robot or an agent in an unseen environment. By reconstructing a depth map of the environment, and marking the boarder between the explored (known) area and unexplored (unknown) area as ``frontiers'', FBE iteratively selects the closest frontier to explore. In addition to being used for constructing depth maps \cite{Verbiest_2015,Leong_2023} and semantic maps \cite{Yu_Kasaei_Cao_2023a,Gomez_Hernandez_Barber_2019} in free-exploration tasks, FBE is also employed in real-world object navigation\cite{Gervet_2022}. FBE has also been adapted in ZSON models with different variations. For example, CLIP on Wheel (CoW) \cite{Gadre_2023} generates text-to-image relevance depth maps based on RGB and depth observations, which is then used to determine the region of interest in FBE \cite{Yamauchi}. ESC \cite{Zhou_2023} and L3MVN \cite{Yu_Kasaei_Cao_2023b} employs FBE by using an LLM to assign scores to each potential frontier. However, the methods of assigning numerical scores to each frontier do not account for the complex interrelations between objects and the environment. Therefore, to solve this problem, our VLTNet aims to use an LLM to incorporate more human-like reasoning in navigation.

\begin{figure*}[h]
    \centering
    \medskip
    \includegraphics[width=1.0\textwidth]{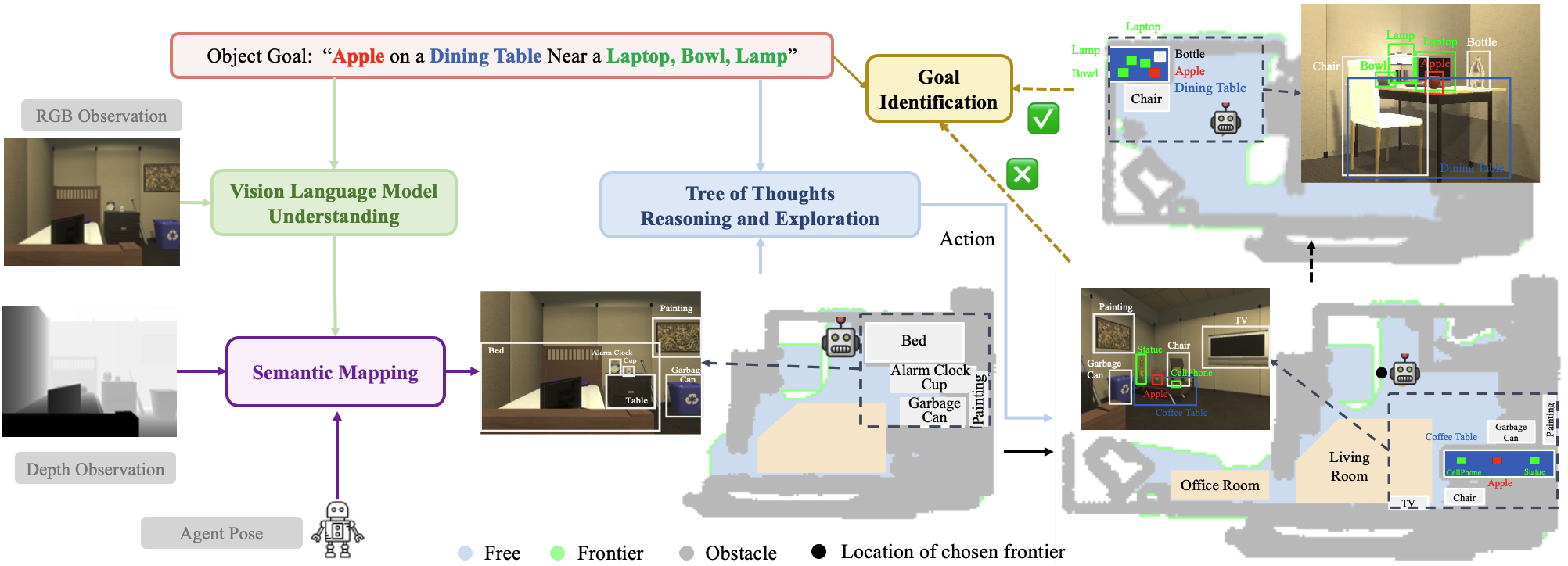}
    \caption{Illustation of our \ourname~framework. During navigation, the \textit{Vision Language Model (VLM) Understanding} module obtains the observed objects by parsing the current RGB observations of an agent. Based on the object locations provided by both the VLM Understanding module and depth observations from the agent, the \textit{Semantic Mapping} module reconstructs a semantic navigation map containing rooms, objects, and frontiers. Conditioned on the navigation instruction and semantic navigation map, the agent then performs common sense reasoning via the \textit{Tree of Thoughts Reasoning and Exploration} module to infer the most probable location of the goal object, and select the corresponding frontier to explore. Upon the VLM Understanding module grounding a candidate object in the same category as the goal object, the \textit{Goal Identification} module further verifies if the candidate object reached by the agent matches the description from the navigation instruction.}
    \label{fig:overview}
\end{figure*}

\paragraph{LLM Reasoning}

Although LLMs have emerged as powerful tools in various domains to understand and generate human-like text \cite{Peters_2018,Devlin_Chang_Lee_Toutanova,Brown_2020}. Since the vanilla LLMs are trained with the aim for natural language processing, they tend to perform poorly in tasks such as arithmetic, common sense, and symbolic reasoning \cite{Rae_2022}. Nonetheless, Wei et al. \cite{Wei_2023} proposed Chain-of-Thought (CoT) prompting to significantly boost LLM's reasoning capability, by instructing them to explicitly output the reasoning process. Building up on the linear progression of thoughts in CoT, Tree-of-Thoughts (ToT) \cite{Yao_2023} proposed a branching reasoning structure that further boosts LLM's reasoning ability, by instructing LLMs to simulate a discussion among several experts on a given question, until reaching a consensus among these simulated experts. In our study, we seek to employ ToT for  decision-making in L-ZSON, which empowers LLMs, \eg, GPT-3.5 \cite{2023gpt3.5} to be able to consider complex interrelations between goal objects and their surroundings, so that LLMs have complete analytical and reasoning autonomy during the frontier selection process.

\section{Methods}
\subsection{Problem Statement}
L-ZSON is designed to validate the capability of an intelligent robot or agent system to navigate to the target or goal objects specified by natural language instructions, without any prior knowledge of the target. In this task, the fundamental components include: 
(1) a natural language instruction $L$, which consists of a sequence of words representing the task to be performed by the agent, encompassing descriptions of the target object, location cues, and directional instructions; 
(2) an environment representation $S_t$, denoting the current state or observation of the agent at time $t$, typically encapsulating the observed information about the environment; and 
(3) a collection of objects within the environment, denoted as $\mathcal{O}$, where each object $o_i \in \mathcal{O}$ is assigned a unique identifier and optionally possesses additional attributes, such as position and appearance. 

The objective of L-ZSON is to generate a sequence of actions $\mathcal{A}$ that guides the agent to navigate within the environment and reach the target object $o^* \in \mathcal{O}$ specified in an instruction $I$, which mathematically represented as: 
\begin{equation}
\mathcal{A}^* = \arg\max_\mathcal{A} P(\mathcal{A} \mid L, S_0, \mathcal{O})
\end{equation}
where $P(\mathcal{A} \mid L, S_0, \mathcal{O})$ represents the probability of a generated action sequence $\mathcal{A}$, given the language instruction $L$, initial state $S_0$, and object set $\mathcal{O}$. The central challenge lies in maximizing the likelihood of selecting the optimal action sequence, enabling the agent to navigate to the target object without prior knowledge or tailored training for that object.

\subsection{\ourname~for L-ZSON}

\subsubsection{Overview}
We present a novel \ourname~tailored for the L-ZSON task, consisting of four core modules as shown in Fig.~\ref{fig:overview}: \textit{Vision Language Model (VLM) Understanding} module, \textit{Semantic Mapping} module, \textit{Tree of Thoughts Reasoning and Exploration} module, and \textit{Goal Identification} module. At each time $t$ during navigation, the \textit{VLM Understanding} module leverages a VLM to perform semantic parsing from the observed RGB image $I_t$, enhancing the model's understanding of the environment semantics. Subsequently, the \textit{Semantic Mapping} module integrates the semantically parsed image $I^{s}_t$ generated from the VLM Understanding module, depth image $D_t$ captured by the agent, and the agent pose $P^a_t$ to construct a more comprehensive semantic map $M_t$, defining objects based on the parsed semantic and spatial relationships. Following that, the \textit{Tree of Thoughts Reasoning and Exploration} module strategically selects a frontier to perform a frontier-based exploration, considering the agent position and the target object information. Lastly, the \textit{Goal Identification} module assesses the alignment of the currently reached object with the goal object specified in the instruction $L$, ensuring navigation consistency. This framework aims to enhance ZSON through a seamless and intelligent integration of scene understanding, semantic mapping, LLM-based frontier selection, and goal object consistency checking, harnessing the power of LLMs equipped with reasoning ability.

\subsubsection{Vision Language Model Understanding}
VLMs excel in semantic understanding, as they have been pre-trained on vast amounts of textual and visual data, %. This pre-training
which enables them to associate texts with the corresponding visual objects, allowing for a deeper comprehension of the content within images.  Specifically, we employ the Grounded Language-Image Pre-training (GLIP) \cite{Li_2022} due to its inherent advantages in grounding language description with visual context. Inspired by ESC \cite{Zhou_2023}, considering both low-level and high-level scene contexts, we define a set of common objects and rooms in an indoor environment as prompts fed into GLIP. We establish multiple prompts, such as the object prompt (\(p_{o}\)) and room prompt (\(p_{r}\)), to query the GLIP model in generating detection results. Here, \(p_{o}\) and \(p_{r}\) correspond to object and room categories, respectively, as represented in natural language. Specifically, at time $t$, we can obtain the detected objects \{$o_{t,i}$\}, rooms \{$r_{t,i}$\} and bounding boxes \{$b^{o}_{t,i}$\} and \{$b^{r}_{t,i}$\} of the objects and rooms from the currently observed image $I_t$: 
\begin{equation}
    \{o_{t,i}, b^{o}_{t,i}, r_{t,i}, b^{r}_{t,i}\}=\mathrm{GLIP}(I_{t}, p_{o},  p_{r}) \in I^s_t 
    \label{eq1}
\end{equation}
where $I^s_t$ is a semantically parsed image.

\subsubsection{Semantic Mapping}
Typically, we need to generate a navigation map that is essential for guiding an agent to make informed decisions during navigation in a complex environment. To achieve this, we utilize the function \(\texttt{Nav\_M}(\cdot)\) to generate the navigation map. Specifically, at time \(t\), we utilize depth information obtained from the agent, along with the agent pose \(P^a_t\), to calculate 3D points from $D_t$. These points are then voxelized into 3D voxels. Subsequently, we project these 3D voxels from the top to produce a 2D navigation map $\mathcal{M}_{nav}$. We formulate the above process as:
\begin{align} 
\mathcal{M}_{nav} &= \texttt{Nav\_M}(D_t, P^a_t).
\label{eq:navmap}
\end{align}
$\mathcal{M}_{nav}$ provides information about the layouts, obstacles, pathways, landmarks, and other relevant details within a specific area. 
Furthermore, we also incorporate the semantic understanding of objects and rooms that are obtained by the VLM Understanding module to generate a semantic navigation map $\mathcal{M}_{sem}$ using $\texttt{Sem\_M}(\cdot)$ function:
\begin{align} 
% \mathcal{M}_{sem} &= \texttt{SematicMapping}(\mathcal M_{nav}, o_{t,i}, b^{o}_{t,i}, r_{t,i}, b^{r}_{t,i})
\mathcal{M}_{sem} &= \texttt{Sem\_M}(\mathcal M_{nav}, \{o_{t,i}, b^{o}_{t,i}, r_{t,i}, b^{r}_{t,i}\}).
\label{eq:sm}
\end{align}
Semantic information, including the types of objects and rooms associated with detected objects in 3D space, is projected onto a 2D plane to create \(\mathcal{M}_{sem}\).
This semantic navigation map $M_t \coloneqq \mathcal{M}_{sem}$ obtained at each time $t$ enables the agent to navigate through the environment with a deeper understanding of the objects and their arrangements, making it more capable of handling complex and dynamic scenarios.
% \hao{(What contains in $\mathcal{M}_{nav}$ and $\mathcal{M}_{sem}$? The function Mapping and SematicMapping is not defined.)}

\subsubsection{Tree-of-Thoughts Reasoning and Exploration}
Due to limitations in the field of agent view or the presence of obstacles, target objects often do not appear within the initial view of an agent. Thus, it is necessary to design an efficient algorithm that enables the agent to explore the environment to swiftly locate the target object quickly. Frontier-based exploration aims at autonomously exploring unknown environments. The core idea is to direct an agent towards the boundaries, known as ``frontiers'', between explored and unexplored areas, ensuring a systematic and efficient exploration. However, traditional frontier-based exploration algorithms \cite{Zhou_2023} usually lead an agent to select the nearest frontier to minimize traversal distance. Given the complexity of certain environments, naively choosing the closest frontier is often not an optimal solution.

To tackle this limitation, we harness the common sense knowledge inherent in LLMs. By analyzing $M_t$, our approach identifies unexplored areas that \textit{are likely proximate to the target object}. Unlike previous methods \cite{Zhou_2023} that rely on Probabilistic Soft Logic (PSL) \cite{bach2017hinge} and craft a bunch of intricate rules to determine the optimal frontier, our approach offers a fresh perspective: \textit{we utilize LLMs to select the frontier that most likely directs to the goal object}. Noting the potential inaccuracies caused by multiple candidate frontiers fed to an LLM in a native way, we integrate the Tree of Thoughts (ToT) mechanism~\cite{Yao_2023} to let the LLM reason about the optimal frontier to select. ToT employs a structured tree-based decision-making process, allowing for organized and systematic exploration, which enhances the model's ability to make informed decisions in complex environments. {Specifically, given a set of frontier candidates $\{f_n\}_{n=1}^N$ return by \cite{Zhou_2023}, we apply the ToT reasoning, as depicted in Algorithm~\ref{alg:tot}~\cite{Yao_2023}, to decide on the optimal frontier for the next move.  To instantiate a ToT, we need to implement four components: thought decomposition, thought generation, state evaluation, and tree search. These components are outlined in the comments of Algorithm~\ref{alg:tot}, which are highlighted in blue.}
% \vspace{-5mm}
\begin{algorithm}[H]
\caption{ToT reasoning($x, m, G, k, V, T, b$)}
\label{alg:tot}
\begin{algorithmic}
\Require Input $x$, an LLM $m$, thought generator $G$ \& size limit $k$, states evaluator $V$, step limit $T$, breadth limit $b$. 
\State $S_0 \gets \{ x \}$ \Comment{\blue{Thought decomposition.}}
\For{$t = 1, \cdots, T$}    \Comment{\blue{Tree search.}}
    \State $S'_t \gets \{ [s, z] \mid s \in S_{t-1}, z_t \in {\color{black}\mathrm{G}}(m, s, k) \}$\Comment{\blue{Thought generation.}}  
    \State $V_t \gets V(m, S'_t)$  \Comment{\blue{Thought evaluation.}}
    \State $S_t \gets \arg \max_{S \subset S'_t, |S| = b} \sum_{s \in S} V_t(s)$ %\Comment{Prune thoughts}
\EndFor \\
\Return $G(m, \arg \max_{s \in S_T} V_T(s), 1)$
\end{algorithmic}
\end{algorithm}
% \vspace{-5mm}
The input $x$ consists of prompt decorator and frontier selection query prompt, and Algorithm \ref{alg:tot} finally returns the selected frontier. We design a prompt decorator or several prompt decorators for each of the above four components to elicit reasoning in LLMs as below.
\begin{itemize}
    \item Thought decomposition: \textit{Imagine ten different experts are answering this question. They will brainstorm the answer step by step, reasoning carefully and taking all facts into consideration.}
    \item Thought generation: \textit{All experts will write down one step of their thinking, then share it with the group. They will each critique their response, and the all the responses of others They will check their answer based on science and the laws of physics. Then all experts will go on to the next step and write down this step of their thinking. They will keep going through steps until they reach their conclusion taking into account the thoughts of the other experts. If at any time they realise that there is a flaw in their logic they will backtrack to where that flaw occurred. If any expert realises they are wrong at any point then they acknowledges this and start another train of thought.}
    \item Thought evaluation: \textit{Each expert will assign a likelihood of their current assertion being correct.}
    \item Tree search: \textit{Continue until the experts agree on a single most likely location.}
\end{itemize}
We append the above ToT prompt decorators with our frontier selection query prompt, \ie, \textit{pick one single location where a laptop is most likely to occur and give a final answer with one single location index}, and the location indices and objects extracted from the semantic navigation map $M_t$ are formatted as: \textit{location \#<i>, located near <room type>, where <\{object1, object2, ...\}> are also found.}. We feed them together into an LLM. The LLM returns a consensus about the most feasible frontier index to the goal object along with a numerical likelihood. The final output from LLM is formated as: \textit{Conclusion, location \#<i> with highest likelihood [\%].} Therefore, our method pinpoints the most promising frontiers, effectively bridging the insights of an LLM with precision in frontier selection, thus enabling more informed and context-aware exploration.

\subsubsection{Goal Identification}
This module determines whether the current object approached by an agent matches the target object specified in an instruction $L$. Our definition of the target object encompasses more intricate spatial and/or appearance descriptions of the object, rather than just object category as previous work \cite{Gadre_2023,Zhou_2023}, such as: ``Alarm clock on a dresser near a desk lamp, bed'' or ``Small, metallic alarm clock''. Thus, an algorithm that merely checks object category, \eg, if the current object is an ``alarm clock'', is insufficient. To make a more informed assessment of whether the scene's context aligns with the target object description, we initially employ a vision language model to interpret the current scene and convert it into a language-based expression. Subsequently, we use a large language model, specifically GPT-3.5~\cite{2023gpt3.5} in our experiments, to analyze the textual descriptions of the target in the instructions $L$ and the object currently observed in the scene. By integrating both textual and visual semantic information, our model achieves a deep semantic understanding of the environment, enhancing the accuracy of aligning scene context with the target description and thereby improving the results of L-ZSON.

\section{Experiments}

\subsection{Environments and Datasets}
We evaluate the performance of our L-ZSON approach based on ToT reasoning on two benchmarks, \ie, PASTURE~\cite{Gadre_2023} and RoboTHOR~\cite{Deitke_2020}.

\paragraph{PASTURE}
Introduced by Gadre \etal in CoW \cite{Gadre_2023}, PASTURE is characterized by its diverse set of environments, each presenting unique navigation challenges. For example, PASTURE introduces categories such as \textit{ uncommon} objects, objects with varying \textit{appearance} complexities, objects placed in intricate \textit{spaces}, and also \textit{hidden} objects strategically obscured from plain sight. Designed mainly for L-ZSON tasks, PASTURE contains 2,520 validation episodes in 15 validation environments with 12 goal object categories. In the PASTURE dataset, agents are tested not only in their navigation skills but also in their adaptability and decision-making ability.

\paragraph{RoboTHOR} 
Introduced by Deitke \etal \cite{Deitke_2020}, offers a platform for ZSON evaluation. Based on real-world indoor settings, RoboTHOR provides precise 3D representations of these environments, creating a more practical and genuine evaluation platform. This benchmark contains a diverse array of objects, set within familiar household and office spaces. It also contains 1,800 validation episodes on 15 validation environments with 12 goal object categories. 
 %While it shares some challenges with PASTURE, like navigating to uncommon objects or concealed objects, ROBOTHOR emphasizes real-world practicality. This requires its ability to not only accurately recognize objects, but also to respond to their ability to move through the space of a real-life scenario.
%Same as PASTURE, the goal objects are mainly considered "small".

\subsection{Metrics}
% \wen{SR and SPL}
% Agents' performance in object navigation tasks is often evaluated with the following metrics:

Following the setting of~\cite{Gadre_2023,Zhou_2023}, we employ the Success Rate (SR) and Success Weighted by Path Length (SWPL) as our evaluation metrics. These metrics not only measure the agent’s ability to reach the goal objects, but also consider the efficiency and reliability of navigation. Specifically, the SR quantifies the proportion of episodes in which the agent successfully navigates to the goal object within maximum steps. Represented in percentage, a higher value suggests superior capability. Although SR provides a measure of success, it does not account for the efficiency of the agent's navigation path. Therefore, the SWPL metric considers both the success of navigation and the optimality of the path taken. It penalizes unnecessary long paths, ensuring that the agent's navigation is both correct and efficient. 

%, evaluating the performance of agents necessitates metrics that not only measure the agent's ability to reach the target but also consider the efficiency and reliability of the navigation. For this purpose, we 
%employ the following metrics to evaluate : 
% 1. Success Rate (SR)

% \begin{equation}
%     SPL = \frac{1}{N} \sum_{i=1}^{N} S_i \times \frac{L_{\text{opt},i}}{\max(L_i, L_{\text{opt},i})}
% \end{equation}
% Where \(S_i\) is the success indicator for episode \(i\), \(L_i\) is the path length of the agent in episode \(i\), and \(L_{\text{opt},i}\) is the shortest possible path length for episode \(i\).

% SR and SPL offer a comprehensive evaluation of the agent's capabilities in object navigation tasks. While SR gives an obvious whether navigation is successful or not, SPL ensures the efficiency of the agent's path, promoting arithmetic that are both effective and optimal.

\subsection{Baselines}
% To ensure a comprehensive evaluation of our ToT reasoning-based exploration approach, we compare its performance against a suite of established baselines, encompassing both traditional methodologies and recent advancements in the realm of Language-driven Zero-Shot Object Navigation (L-ZSON).

Our VLTNet is evaluated against the following state-of-the-art models for both ZSON and L-ZSON tasks.

\noindent \textbf{CoW} \cite{Gadre_2023}: CoW targets both ZSON and L-ZSON tasks, using CLIP to consistently update a top-down map with image-to-goal relevance. Variants of CoW with different CLIP-like localization modules were also included: \textbf{CLIP-Ref~\cite{Gadre_2023}, CLIP-Patch~\cite{Gadre_2023}, CLIP-Grad~\cite{Gadre_2023}, MDETR~\cite{Kamath_2021}, OWL~\cite{Minderer_2022}.}

\noindent \textbf{ESC}\cite{Zhou_2023}: ESC utilizes GLIP for object detection to facilitate scene understanding and common sense reasoning. ESC also incorporates soft logic predicates to ensure optimal path and navigation decisions.

% \textbf{Frontier-based Exploration (with Soft Logic Predicates):} Building upon the foundational principles of FBE, the ESC method introduces an innovative twist by incorporating soft logic predicates. These predicates, derived from common sense knowledge, directing the navigation agent towards optimal paths and decisions. This fusion of heuristic exploration with common sense reasoning offers a robust navigation strategy, adept at handling both conventional and edge-case scenarios.

\subsection{Results}
% \usepackage{multirow}
% \vspace{3em}

% \medskip

% \usepackage{multirow}
\begin{table*}[tb]
\centering
\footnotesize
\caption{Quantitative results on the PASTURE\cite{Gadre_2023} and RoboTHOR\cite{AI} benchmarks are provided, comparing our VLTNet model with six CoW (CLIP on Wheel) variants designed for L-ZSON tasks, while ESC is exclusively used for ZSON tasks. Abbreviations used include Unc. for Uncommon, App. for Appearance, dist. for distract, and Hid. for Hidden. The best results are highlighted in red bold, while the second-best results are highlighted in blue bold.}
\begin{tabular}{c|ccccccccc|cc}
\toprule
 &
  \multicolumn{9}{c|}{\textbf{PASTURE}} &
  \multicolumn{2}{c}{\textbf{RoboTHOR}} \\ \cmidrule{2-12} 
 &
  \textbf{Unc.} &
  \textbf{App.} &
  \textbf{Space} &
  \textbf{\begin{tabular}[c]{@{}c@{}}App.\\ dist.\end{tabular}} &
  \textbf{\begin{tabular}[c]{@{}c@{}}Space\\ dist.\end{tabular}} &
  \textbf{Hid.} &
  \textbf{\begin{tabular}[c]{@{}c@{}}Hid.\\ dist.\end{tabular}} &
  \multicolumn{2}{c|}{\textbf{Avg.}} & \multicolumn{2}{c|}{\textbf{Avg.}}
   \\
\multirow{-3}{*}{\textbf{Models}} &
  SR &
  SR &
  SR &
  SR &
  SR &
  SR &
  SR &
  SWPL &
  SR &
  SWPL &
  SR \\ \midrule
CLIP-Ref. &
  {3.6} &
  {2.8} &
  {2.8} &
  {3.1} &
  {3.3} &
  {4.7} &
  {5.0} &
  {1.7} &
  {2.5} &
  {2.4} &
  {2.7} \\
CLIP-Patch &
  18.1 &
  13.3 &
  13.3 &
  10.8 &
  10.8 &
  17.5 &
  {\color[HTML]{3531FF} \textbf{17.8}} &
  9.0 &
  14.2 &
  10.6 &
  20.3 \\
CLIP-Grad. &
  16.1 &
  11.9 &
  11.7 &
  9.7 &
  10.3 &
  14.4 &
  16.1 &
  9.2 &
  12.9 &
  9.7 &
  15.2 \\
MDETR &
  3.1 &
  7.2 &
  5.0 &
  7.2 &
  4.7 &
  8.1 &
  8.9 &
  5.4 &
  6.3 &
  8.4 &
  9.9 \\
OWL &
  32.8 &
  {\color[HTML]{3531FF} \textbf{26.9}} &
  {\color[HTML]{3531FF} \textbf{19.4}} &
  {\color[HTML]{3531FF} \textbf{19.4}} &
  {\color[HTML]{3531FF} \textbf{16.1}} &
  {\color[HTML]{3531FF} \textbf{19.2}} &
  15.8 &
  {\color[HTML]{3531FF} \textbf{12.6}} &
  {\color[HTML]{3531FF} \textbf{21.1}} &
  17.2 &
  27.5 \\
ESC &
  {\color[HTML]{3531FF} \textbf{35.5}} &
  - &
  - &
  - &
  - &
  - &
  - &
  - &
  - &
  {\color[HTML]{FE0000} \textbf{22.2}} &
  {\color[HTML]{FE0000} \textbf{38.1}} \\ \midrule
VLTNet &
  {\color[HTML]{FE0000} \textbf{36.9}} &
  {\color[HTML]{FE0000} \textbf{35.0}} &
  {\color[HTML]{FE0000} \textbf{33.3}} &
  {\color[HTML]{FE0000} \textbf{21.9}} &
  {\color[HTML]{FE0000} \textbf{21.7}} &
  {\color[HTML]{FE0000} \textbf{22.8}} &
  {\color[HTML]{FE0000} \textbf{26.4}} &
  {\color[HTML]{FE0000} \textbf{14.0}} &
  {\color[HTML]{FE0000} \textbf{28.2}} &
  17.1 &
  {\color[HTML]{3531FF} \textbf{33.2}} \\ 
  \bottomrule
\end{tabular}
\label{table:res}
\end{table*}

\begin{figure}
    \centering
    \medskip
    \includegraphics[width=1\linewidth]{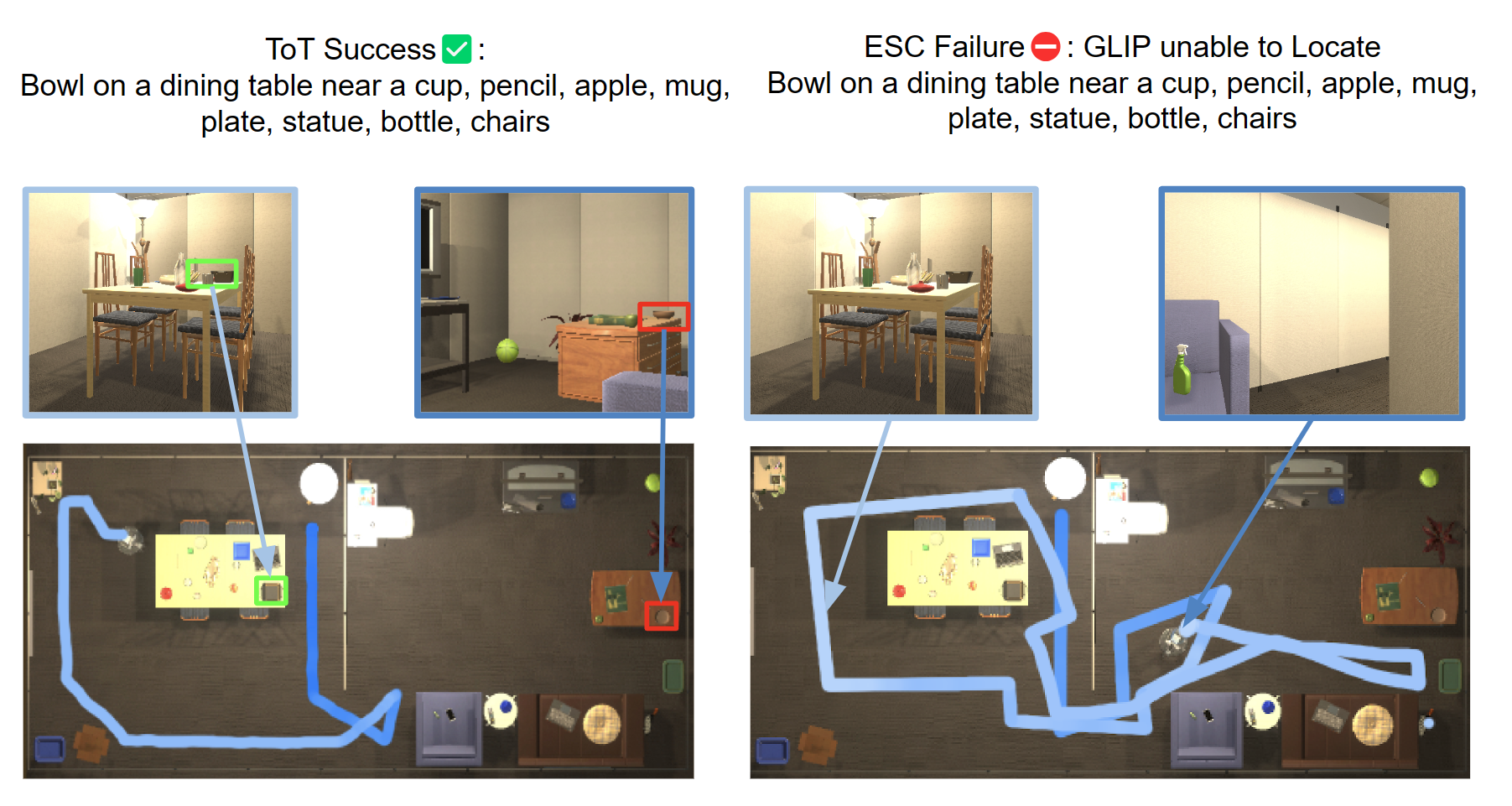}
    % \vspace{-8mm}
    \caption{Visualizing egocentric trajectories of VLTNet and ESC navigation process when given a spatial goal instruction. Color indicates trajectory progress, where blue indicating trajectory start and white indicating trajectory end. The goal objects are boxed in green, while distractors are boxed in red.}
    \label{fig3}
\end{figure}

% \begin{figure}
%     \centering
%     \includegraphics[width=1\linewidth]{Spatial_Comparison_With_Cow.png}
%     \caption{Visualizing trajectories of ToT-Prompted Navigation and CoW Navigation when navigating with spatial goal description. Frames are egocentric views. Color indicates trajectory progress, where blue indicating trajectory start and white indicating trajectory end. Target objects are boxed in green, while distractor objects are boxed in red}
%     \label{fig2}
% \end{figure}

Our experiments was designed rigorously to assess the efficacy of our proposed VLTNet for ZSON and L-ZSON tasks. We juxtaposed our method with the state-of-the-art approaches and the results are shown in Table \ref{table:res}.

On the PASTURE dataset, 
% tailored for gauging navigation algorithms against rare objects and both spatial and appearance-based targets, 
our VLTNet model consistently surpassed competing models across all metrics. Notably, within the \textit{Appearance} category, our VLTNet model achieves a noteworthy success rate of 35.0\%. In contrast, the OWL has an SR of 26.9\%. 
% This underscores our model's advanced ability to discern and navigate using subtle visual cues using a combination of VLM Understanding and Goal Identification Module, enhancing its aptitude in navigating based on object appearances.
%
% Our experiments were meticulously designed to evaluate the performance of our proposed ToT method in the context of both ZSON and LD-ZSN tasks. We benchmarked our approach against state-of-the-art (SOTA) methods, including those presented in the CoW and ESC studies.
%
% In the PASTURE dataset, which is designed to evaluate navigation algorithms in the presence of uncommon objects, spatial-based and appearance-based targets, our VLTNet model, backed by the integration of GLIP, consistently outperformed other models across all these categories. Specifically, in the "Appearance" category, VLTNet model achieved an impressive success rate (SR) of 35.0\%. Comparatively, the CoW model, secured an SR of 26.9\%. This is a testament to our model's capability to process and understand intricate visual cues, enabling the model to navigate based on object appearances more effectively.
%
Similarly, in the \textit{Spatial} category, the SR of our VLTNet model is 33.3\%, outperforming OWL model's 19.4\%. This demonstrates our model's capability in understanding spatial relationships and interpreting complex object descriptions using the Tree of Thoughts Reasoning and Exploration module. As shown in Fig. \ref{fig3}, our model successfully leverages an LLM to extract the candidate frontier of ``bowl'' and then the Goal Identification module verifies that the bowl aligns with the spatial cues in an instruction. Conversely, the ESC model is unable to locate the goal object even if the agent was facing the target.
% , mostly because ESC's object localizer (GLIP) suffers from severe performance degradation when given raw complex object prompts.  Nonetheless, 
Also, it is essential to note that ESC is only designed for ZSON tasks and thus can only accept a single object category instruction and cannot directly handle object descriptions using natural language.
% That is why all metrics that ensembles L-ZSON tasks are left blank for ESC.

On the RoboTHOR dataset, the ESC model, tailored for RoboTHOR, secures an SR of 38.1\%. Our VLTNet continues its commendable performance by achieving an SR of 33.2\% and an SWPL of 17.1\%, which outperforms CoW that secures an SR of 27.5\%. This further proves that our VLTNet navigation model has a competitive performance compared to the state-of-the-art methods.

\subsection{Ablation Study}

\paragraph{\textbf{The effect of ToT Reasoning and Exploration module.}}
% As previously mentioned, our VLTNet's ToT Reasoning and Exploration module aims to incorporate Tree-of-Thoughts reasoning during the Frontier Selection process. Given LLM's nature of lacking coherence in complex common sense reasoning, this module was able to further elicit LLM's common sense reasoning capability and to make optimal exploration decisions under complex environments.
%
To evaluate the efficacy of Tree of Thoughts Reasoning and Exploration module, we conducted a comparative analysis with two models on the PASTURE LONGTAIL dataset \cite{Gadre_2023}, consisting of 12 uncommon object goals. All models employ GPT-3.5, differing only in their input prompts. The first model is guided to directly select a frontier from all the available candidates, devoid of any explicit directive for reasoning. The second model uses ToT input prompts, which requires a deliberation between ten experts to articulate their reasoning and collectively determine a frontier to select for exploration. As evidenced by Table \ref{table:ab_1}, the model that uses ToT prompts for frontier selection exhibits a marked superiority over the model without ToT prompts. This underscores the efficacy of ToT prompting in facilitating the selection of frontier that are closer to the goal object. 

% \begin{table}[htbp]
%   \caption{Ablation Study} 
%   \subfloat[Performance between different prompting in ToT Reasoning and Exploration Module]{%
%     \scalebox{0.75}{\begin{tabular}{lllll}
% \hline
%                             &                   & \multicolumn{3}{l}{Pasture Uncom.}                                \\ \cline{3-5} 

% \multicolumn{2}{l}{Reasoning Prompt}            & SPL                   & \multicolumn{2}{l}{SR}                    \\ \hline
% \multicolumn{2}{l}{W/o ToT Prompt}              & 12.4                  & \multicolumn{2}{l}{29.8}                  \\
% \multirow{ToT Prompt} & \multirow{} & \multirow{\textbf{16.6}} & \multirow{\textbf{36.9}} & \multirow{}            \\\hline                   
% \end{tabular}}
%    %
%   }
%    \hspace{.5cm}
%  \subfloat[Performance between different models in Goal Identification Module]{%
%    \scalebox{0.75}{\begin{tabular}{lllll}
% \hline
% \multicolumn{2}{l}{}                  & \multicolumn{3}{l}{Pasture Uncom.}                     \\ \cline{3-5} 
% \multicolumn{2}{l}{Validation Methods} & SPL                & \multicolumn{2}{l}{SR}            \\ \hline
% \multicolumn{2}{l}{GLIP}              & 5.9                & \multicolumn{2}{l}{12.6}          \\
% \multicolumn{2}{l}{ViLT}              & 8.7                & \multicolumn{2}{l}{18.3}          \\
% \multicolumn{2}{l}{GPT-3.5}           & \textbf{9.3}       & \multicolumn{2}{l}{\textbf{21.7}} \\ \hline
% \end{tabular}}
%     % \hspace{.5cm}%
%   }
% \label{table:2}
% \end{table}

\begin{table}[ht]
\centering
\vspace{-4mm}
\begin{minipage}[b]{0.45\linewidth}
\centering
\caption{Performance between different prompting in ToT Reasoning and Exploration module on Pasture Uncom. split.}
\label{table:ab_1}
\begin{tabular}{lcc}
\toprule
Reasoning Prompt & SWPL & SR \\
\midrule
W/o ToT prompts & 12.4 & 29.8 \\
ToT prompts & \textbf{16.6} & \textbf{36.9} \\
\bottomrule
\end{tabular}
\end{minipage}
\hfill
\begin{minipage}[b]{0.45\linewidth}
\centering
\caption{Performance between different models in Goal Identification module on Pasture Space dist. split.}
\label{table:ab_2}
\begin{tabular}{lcc}
\toprule
Module & SWPL & SR \\
\midrule
GLIP & 5.9 & 12.6 \\
ViLT & 8.7 & 18.3 \\
GPT-3.5 & \textbf{9.3} & \textbf{21.7} \\
\bottomrule
\end{tabular}
\end{minipage}
\vspace{-4mm}
\end{table}

\paragraph{\textbf{Comparison of different models for Goal Identification module.}}

To prove the robustness of using an LLM in the Goal Identification module, we tested this module using GPT-3.5 along with two other VLM models: VILT \cite{Kim_Son_Kim_2021} for visual question answering and GLIP \cite{Li_2022} for object grounding. 
% For VLM-based validations, upon potential object goal localization by VLM Understanding Module, the current frame is fed into the VLM, alongside the original descriptive object prompt. For the GPT-3.5 model, detected object captions are integrated into a prompt describing the scene, with GPT-3.5 subsequently assessing the congruence between the present scene description and the provided spatial cue. 
All three models are evaluated on the PASTURE Space dataset \cite{Gadre_2023}, in which target objects are embedded in spatially descriptive prompts. Table \ref{table:ab_2} illustrates that GLIP faces challenges in grounding objects when presented with intricate spatial cues. When the current frame is isolated and processed through VILT, there is a marginal improvement in object identification based on spatial hints. However, the most effective method for validating the goal object in accordance with a spatial prompt is GPT-3.5, by determining the congruence between objects present in the current scene and the provided spatial cues in an instruction.

\section{CONCLUSIONS}

In this paper, we introduce a VLTNet model, which harnesses both visual language modeling and ToT reasoning for L-ZSON task. We innovatively integrated the Tree of Thoughts reasoning framework, enriching the decision-making process with its nuanced multi-path reasoning capabilities. This empowers the model to make informed decisions during a frontier selection process in language-instructed navigation. The results on the PASTURE and RoboTHOR benchmarks demonstrate that our VLTNet excels in handling complex L-ZSON tasks that demand intricate understanding and interpretation of natural language instructions and environments.

% \subsubsection{Acknowledgements} Please place your acknowledgments at the end of the paper, preceded by an unnumbered run-in heading (i.e. 3rd-level heading).

% \newpage
\bibliographystyle{splncs04}
\bibliography{egbib}

\begin{thebibliography}{10}
\providecommand{\url}[1]{\texttt{#1}}
\providecommand{\urlprefix}{URL }
\providecommand{\doi}[1]{https://doi.org/#1}

\bibitem{AI}
AI, A.I.f.: Key features, \url{https://ai2thor.allenai.org/robothor/}

\bibitem{Halah_2022}
Al-Halah, Z., Ramakrishnan, S.K., Grauman, K.: Zero experience required: Plug \& play modular transfer learning for semantic visual navigation (Apr 2022), \url{https://arxiv.org/abs/2202.02440}

\bibitem{bach2017hinge}
Bach, S.H., Broecheler, M., Huang, B., Getoor, L.: Hinge-loss markov random fields and probabilistic soft logic. Journal of Machine Learning Research  \textbf{18}(109),  1--67 (2017)

\bibitem{Brown_2020}
Brown, T., Mann, B., Ryder, N., Subbiah, M., Kaplan, J.D., Dhariwal, P., Neelakantan, A., Shyam, P., Sastry, G., Askell, A., et~al.: Language models are few-shot learners (Jan 2020), \url{https://proceedings.neurips.cc/paper_files/paper/2020/hash/1457c0d6bfcb4967418bfb8ac142f64a-Abstract.html}

\bibitem{Chang_Gupta_Gupta_2020}
Chang, M., Gupta, A., Gupta, S.: Semantic visual navigation by watching youtube videos (Jan 2020), \url{https://proceedings.neurips.cc/paper/2020/hash/2cd4e8a2ce081c3d7c32c3cde4312ef7-Abstract.html}

\bibitem{Chaplot}
Chaplot, D.S., Gandhi, D., Gupta, S., Gupta, A., Salakhutdinov, R.: Learning to explore using active neural slam, \url{https://iclr.cc/virtual_2020/poster_HklXn1BKDH.html}

\bibitem{Chatto_2021}
Chattopadhyay, P., Hoffman, J., Mottaghi, R., Kembhavi, A.: Robustnav: Towards benchmarking robustness in embodied navigation (Jun 2021), \url{https://arxiv.org/abs/2106.04531}

\bibitem{Chen_2022}
Chen, P., Ji, D., Lin, K., Zeng, R., Li, T.H., Tan, M., Gan, C.: Weakly-supervised multi-granularity map learning for vision-and-language navigation (Oct 2022), \url{https://arxiv.org/abs/2210.07506}

\bibitem{Deitke_2020}
Deitke, M., Han, W., Herrasti, A., Kembhavi, A., Kolve, E., Mottaghi, R., Salvador, J., Schwenk, D., VanderBilt, E., Wallingford, M., et~al.: Robothor: An open simulation-to-real embodied ai platform (Apr 2020), \url{https://arxiv.org/abs/2004.06799}

\bibitem{Deitke_2022}
Deitke, M., VanderBilt, E., Herrasti, A., Weihs, L., Ehsani, K., Salvador, J., Han, W., Kolve, E., Kembhavi, A., Mottaghi, R.: Procthor: Large-scale embodied ai using procedural generation (Dec 2022)

\bibitem{Devlin_Chang_Lee_Toutanova}
Devlin, J., Chang, M.W., Lee, K., Toutanova, K.: Bert: Pre-training of deep bidirectional transformers for language understanding, \url{https://aclanthology.org/N19-1423/}

\bibitem{Dorbala_2022}
Dorbala, V.S., Sigurdsson, G., Piramuthu, R., Thomason, J., Sukhatme, G.S.: Clip-nav: Using clip for zero-shot vision-and-language navigation (Nov 2022), \url{https://arxiv.org/abs/2211.16649}

\bibitem{Gadre_2023}
Gadre, S.Y., Wortsman, M., Ilharco, G., Schmidt, L., Song, S.: Cows on pasture: Baselines and benchmarks for language-driven zero-shot object navigation. 2023 IEEE/CVF Conference on Computer Vision and Pattern Recognition  (2023). \doi{10.1109/cvpr52729.2023.02219}

\bibitem{Gervet_2022}
Gervet, T., Chintala, S., Batra, D., Malik, J., Chaplot, D.S.: Navigating to objects in the real world (Dec 2022), \url{https://arxiv.org/abs/2212.00922}

\bibitem{Gomez_Hernandez_Barber_2019}
Gomez, C., Hernandez, A.C., Barber, R.: Topological frontier-based exploration and map-building using semantic information (Oct 2019), \url{https://www.mdpi.com/1424-8220/19/20/4595}

\bibitem{He_2017}
He, K., Gkioxari, G., Dollar, P., Girshick, R.: Mask r-cnn. IEEE International Conference on Computer Vision  (2017). \doi{10.1109/iccv.2017.322}

\bibitem{huang2023visual}
Huang, C., Mees, O., Zeng, A., Burgard, W.: Visual language maps for robot navigation. In: IEEE International Conference on Robotics and Automation. pp. 10608--10615. IEEE (2023)

\bibitem{Kamath_2021}
Kamath, A., Singh, M., LeCun, Y., Synnaeve, G., Misra, I., Carion, N.: Mdetr - modulated detection for end-to-end multi-modal understanding. IEEE/CVF International Conference on Computer Vision  (2021)

\bibitem{Khandelwal_2022}
Khandelwal, A., Weihs, L., Mottaghi, R., Kembhavi, A.: Simple but effective: Clip embeddings for embodied ai. IEEE/CVF Conference on Computer Vision and Pattern Recognition  (2022). \doi{10.1109/cvpr52688.2022.01441}

\bibitem{Kim_Son_Kim_2021}
Kim, W., Son, B., Kim, I.: Vilt: Vision-and-language transformer without convolution or region supervision (Jun 2021), \url{https://arxiv.org/abs/2102.03334}

\bibitem{Leong_2023}
Leong, K.: Reinforcement learning with frontier-based exploration via autonomous environment (Jul 2023), \url{https://arxiv.org/abs/2307.07296}

\bibitem{Li_2022}
Li, L.H., Zhang, P., Zhang, H., Yang, J., Li, C., Zhong, Y., Wang, L., Yuan, L., Zhang, L., Hwang, J.N., et~al.: Grounded language-image pre-training. IEEE/CVF Conference on Computer Vision and Pattern Recognition  (2022). \doi{10.1109/cvpr52688.2022.01069}

\bibitem{Majumdar_2022}
Majumdar, A., Aggarwal, G., Devnani, B., Hoffman, J., Batra, D.: Zson: Zero-shot object-goal navigation using multimodal goal embeddings (Jun 2022)

\bibitem{Maksymets_Cartillier_Gokaslan_Wijmans_Galuba_Lee_Batra_2021}
Maksymets, O., Cartillier, V., Gokaslan, A., Wijmans, E., Galuba, W., Lee, S., Batra, D.: Thda: Treasure hunt data augmentation for semantic navigation. IEEE/CVF International Conference on Computer Vision  (2021). \doi{10.1109/iccv48922.2021.01509}

\bibitem{Mezghan_2022}
Mezghan, L., Sukhbaatar, S., Lavril, T., Maksymets, O., Batra, D., Bojanowski, P., Alahari, K.: Memory-augmented reinforcement learning for image-goal navigation. IEEE/RSJ International Conference on Intelligent Robots and Systems  (2022). \doi{10.1109/iros47612.2022.9981090}

\bibitem{Min_2023}
Min, S.Y., Chaplot, D.S., Ravikumar, P.K., Bisk, Y., Salakhutdinov, R.: Film: Following instructions in language with modular methods (Sep 2023), \url{https://openreview.net/forum?id=qI4542Y2s1D}

\bibitem{Minderer_2022}
Minderer, M., Gritsenko, A., Stone, A., Neumann, M., Weissenborn, D., Dosovitskiy, A., Mahendran, A., Arnab, A., Dehghani, M., Shen, Z., et~al.: Simple open-vocabulary object detection with vision transformers (Jul 2022), \url{https://arxiv.org/abs/2205.06230}

\bibitem{2023gpt3.5}
OpenAI: Gpt-3.5 technical report  (2023)

\bibitem{Park_Yoon_023}
Park, J., Yoon, T., Hong, J., Yu, Y., Pan, M., Choi, S.: Zero-shot active visual search (zavis): Intelligent object search for robotic assistants. IEEE International Conference on Robotics and Automation  (2023). \doi{10.1109/icra48891.2023.10161345}

\bibitem{Peters_2018}
Peters, M.E., Neumann, M., Iyyer, M., Gardner, M., Clark, C., Lee, K., Zettlemoyer, L.: Deep contextualized word representations (Mar 2018), \url{https://arxiv.org/abs/1802.05365}

\bibitem{Radford_2021}
Radford, A., Kim, J.W., Hallacy, C., Ramesh, A., Goh, G., Agarwal, S., Sastry, G., Askell, A., Mishkin, P., Clark, J., et~al.: Learning transferable visual models from natural language supervision (Feb 2021), \url{https://arxiv.org/abs/2103.00020}

\bibitem{Rae_2022}
Rae, J.W., Borgeaud, S., Cai, T., Millican, K., Hoffmann, J., Song, F., Aslanides, J., Henderson, S., Ring, R., Young, S., et~al.: Scaling language models: Methods, analysis \& insights from training gopher (Jan 2022), \url{https://arxiv.org/abs/2112.11446}

\bibitem{Ramrakhya_2022}
Ramrakhya, R., Undersander, E., Batra, D., Das, A.: Habitat-web: Learning embodied object-search strategies from human demonstrations at scale. IEEE/CVF Conference on Computer Vision and Pattern Recognition  (2022). \doi{10.1109/cvpr52688.2022.00511}

\bibitem{Verbiest_2015}
Verbiest, K., Berrabah, S.A., Colon, E.: Autonomous frontier based exploration for mobile robots (Jan 2015), \url{https://link.springer.com/chapter/10.1007/978-3-319-22873-0_1}

\bibitem{Wei_2023}
Wei, J., Wang, X., Schuurmans, D., Bosma, M., Ichter, B., Xia, F., Chi, E., Le, Q., Zhou, D.: Chain-of-thought prompting elicits reasoning in large language models (Jan 2023), \url{https://arxiv.org/abs/2201.11903}

\bibitem{Yamauchi}
Yamauchi, B.: A frontier-based approach for autonomous exploration. Proceedings IEEE International Symposium on Computational Intelligence in Robotics and Automation . \doi{10.1109/cira.1997.613851}

\bibitem{Yao_2023}
Yao, S., Yu, D., Zhao, J., Shafran, I., Griffiths, T.L., Cao, Y., Narasimhan, K.: Tree of thoughts: Deliberate problem solving with large language models (May 2023), \url{https://arxiv.org/abs/2305.10601}

\bibitem{Ye_Batra_Das_Wijmans_2021}
Ye, J., Batra, D., Das, A., Wijmans, E.: Auxiliary tasks and exploration enable objectgoal navigation. IEEE/CVF International Conference on Computer Vision  (2021). \doi{10.1109/iccv48922.2021.01581}

\bibitem{Yu_Kasaei_Cao_2023a}
Yu, B., Kasaei, H., Cao, M.: Frontier semantic exploration for visual target navigation. IEEE International Conference on Robotics and Automation  (2023). \doi{10.1109/icra48891.2023.10161059}

\bibitem{Yu_Kasaei_Cao_2023b}
Yu, B., Kasaei, H., Cao, M.: L3mvn: Leveraging large language models for visual target navigation (Apr 2023), \url{https://arxiv.org/abs/2304.05501}

\bibitem{Zhao_2023}
Zhao, Q., Zhang, L., He, B., Qiao, H., Liu, Z.: Zero-shot object goal visual navigation. IEEE International Conference on Robotics and Automation  (2023). \doi{10.1109/icra48891.2023.10161289}

\bibitem{Zheng_2022}
Zheng, K., Zhou, K., Gu, J., Fan, Y., Wang, J., Di, Z., He, X., Wang, X.E.: Jarvis: A neuro-symbolic commonsense reasoning framework for conversational embodied agents (Sep 2022), \url{https://arxiv.org/abs/2208.13266}

\bibitem{Zhou_2023}
Zhou, K., Zheng, K., Pryor, C., Shen, Y., Jin, H., Getoor, L., Wang, X.E.: Esc: Exploration with soft commonsense constraints for zero-shot object navigation (Jul 2023), \url{https://arxiv.org/abs/2301.13166}

\bibitem{Zhu_2017}
Zhu, Y., Mottaghi, R., Kolve, E., Lim, J.J., Gupta, A., Fei-Fei, L., Farhadi, A.: Target-driven visual navigation in indoor scenes using deep reinforcement learning. IEEE International Conference on Robotics and Automation  (2017). \doi{10.1109/icra.2017.7989381}

\end{thebibliography}

\end{document}